\documentclass{article}

\usepackage[utf8]{inputenc} 
\usepackage[T1]{fontenc}    
\usepackage{hyperref}       
\usepackage{url}            
\usepackage{booktabs}       
\usepackage{amsfonts}       
\usepackage{nicefrac}       
\usepackage{microtype}      
\usepackage{xcolor}         
\usepackage[preprint]{neurips_2022}
\usepackage{algpseudocode}
\usepackage{algorithm}
\usepackage{graphicx}

\title{Generate synthetic samples from tabular data}

\author{%
David Banh$^{1}$ \quad Alan Huang$^{2}$ \\
$^1$AskExplain \quad $^2$University of Queensland \\
\texttt{david.b@askexplain.com}  \ \ \ \texttt{alan.huang@uq.edu.au}
}

\begin{document}

\maketitle

\begin{abstract}

Generating new samples from data sets can mitigate extra expensive operations, increased invasive procedures, and mitigate privacy issues. 
These novel samples that are statistically robust can be used as a temporary and intermediate replacement when privacy is a concern. This method can enable better data sharing practices without problems relating to identification issues or biases that are flaws for an adversarial attack. 

\end{abstract}

Keywords:
data sampling, balanced sampling, data sharing, data access, data privacy, latent samples

GitHub:
https://github.com/AskExplain/synthetic\_sampling

\section{Introduction}

Mitigating the restrictions caused by privacy and safety concerns for individuals can be solved by generating computationally new samples without identifiers. This can resolve the many and various trade-offs of enforced privacy on data [1], [2], [3]. Here, generating novel samples from observed data is shown to unexpectedly improve accuracy, precision, statistical robustness and unbiasedness, due to the benefits of generating large and balanced sample sizes.

The technique known as linear encoding shares many fundamental aspects that investigate structure, balance and dimensionality. Structure relates to the concepts of graphical structures including Bayesian networks and complex or random graphs, such as with model-based sampling [4]. The idea of balanced data ensures the (known or unknown) sample classes are proportionally representative such as with Synthetic Minority Over-sampling Technique or SMOTE [5]. Dimensionality involves viewing the data by summarising into core components and extending the space like with [6].

\section{Model}

\subsection{Linear Encoding}

$$ \alpha_Y Y \beta_y = \alpha_X X \beta_X = \alpha_Z Z \beta_Z $$

The datasets $Y$, $X$ and $Z$ can be of any size, from matrix to tensor.

$$ Y = \alpha_Y^T C_Y \beta_Y^T + i_Y + \epsilon_Y $$
$$ X = \alpha_X^T C_X \beta_X^T + i_X + \epsilon_X $$
$$ Z = \alpha_Z^T C_Z \beta_Z^T + i_Z + \epsilon_Z $$

Multiple modalities of data can be projected or "encoded" into the same subspace. 

The projections $\alpha$ and $\beta$ transform the datasets $Y$, $X$, and $Z$ into a latent dimensional space. The parameters $\alpha$ and $\beta$ also compose or factorise the data.

A common projection via transformation to a shared space is constructed through the latent code $C$. The latent code can be fixed at $C_Y = C_X = C_Z$, and is constrained across all datasets. This restricts all parameters learned to the same space, and projects all models into a common plane.

Notice the similarities to Singular Value Decomposition (SVD). To make extensions to SVD an intercept $i_{j}$ can be added with residuals $\epsilon$ from a distribution (e.g. Gaussian, Poisson, Negative Binomial). Furthermore, an algebraic extension is also shown via the shared space through the common latent code  $C = C_Y = C_X = C_Z$.

\subsection{Coordinate descent updates for learning parameters}

The coordinate descent update to estimate the parameters is very simple and straightforward. It iterates through multiple steps outlined by Algorithm 1 until convergence. The full step loops over multiple datasets (and runs across a complete set, full list, or tuple of datasets):

\begin{algorithm}
    \caption{Linear Encoding}\label{euclid} 
    \textbf{Input}: $D_L$ Dataset for each modality $L$  \\ 
    \textbf{Output}: $\alpha_L$ sample parameters, $\beta_L$ feature parameters, $C$ code \\ 

            For each object in the list of data sets

                \Comment{Update sample parameters} \\ 
                $ \alpha_L = D_L^T (C \beta^T_L)^T ( (C \beta^T_L) (C \beta^T_L)^T )^{-1}$  
                
                \Comment{Update feature parameters} \\ 
                $ \beta_L = ((\alpha^T_L C)^T (\alpha^T_L C))^{-1} (\alpha^T_L C)^T D_L $ 
                
                \Comment{Update Code} \\ 
                $ C = (\alpha_L^T \alpha_L)^{-1} \alpha^T_L D_L \beta^T_L ( \beta_L \beta^T_L)^{-1} $

\end{algorithm}

\section{Generative Sampling}

Following, the $\alpha$ model is treated as a latent sampling structure that considers how samples are related. Based on this information, a Gaussian mixture model is applied to learn the distribution of the latent space $i$:

$$ A_{i,g} \sim \pi_{g} N(\mu_{g}, D_{g}) $$

However, there is a more potent method based on Lindenbaum et al work [6] with recent literature stating the local manifold of the data can be sampled with greater representation in structure. The work is based on sampling local Gaussian covariance structures from the manifold learned using K-nearest neighbours. 

For example, identify a large number of centroids using K-nearest neighbours. Following, take a centroid $C$ and the nearest points $K$. Extract two parameters from nearest points $K$ that define the local manifold at that centroid $C$: $\eta$ the mean of the sampled points, and $\Omega$ the covariance of the sampled points. Following, generate random Gaussian variables $s$ based on these details for each centroid $C$, giving the parameter $$A_{i,C_{s}} \sim N(\eta_C,\Omega_C) $$ where $C_{s}$ is the number of sampled points $s$ per centroid $C$. One can imagine this process from Lindenbaum et al [6] samples a "Gaussian mixture model" at each local centroid.

Finally, the response $Y$ and the covariates $X$ are projected from the latent sample space $\alpha$ and the encoding-based latent sample space $A$ via the expression:

$$ A_{i,g}^T (\alpha \alpha^T)^{-1} \alpha X = A_{i,g}^T (\alpha \alpha^T)^{-1} \alpha \alpha^T C \beta_X = A_{i,g}^T C \beta_{X} = \hat{X} $$
$$ A_{i,g}^T (\alpha \alpha^T)^{-1} \alpha Y = A_{i,g}^T (\alpha \alpha^T)^{-1} \alpha \alpha^T C \beta_Y = A_{i,g}^T C \beta_{Y} = \hat{Y} $$
 
\section{Results}

\subsection{Regression on real and synthetic datasets}

The test here is to compare the regression performance on real datasets against synthetic datasets generated using the real dataset as a foundation. In all cases, the synthetic datasets outperformed in terms of Mean-Absolute Deviation (M.A.D) on all 10 datasets in a statistically significant manner (P-value comparing M.A.D for a standard T-test). The regression technique used is glmnet [7].

\begin{center}
\begin{tabular}{ |c||c| } 
 \hline
    Dataset & P-value \\
     \hline \hline
    Wine & 1.32e-59 \\
    Oliveoil & 1.76e-42 \\
    Glass & 2.53e-13 \\
    Thyroid & 1.96e-37 \\
    WDBC & 2.64e-64 \\
    PimaIndiansDiabetes & 3.18e-63 \\
    BostonHousing & 1.52e-44 \\
    Ionosphere & 9.74e-170 \\
    Shuttle & 4.58e-14 \\
    Satellite & 1.59e-38 \\
 \hline
\end{tabular}
\end{center}

The ten datasets are sourced from R library packages (pdfCluster, mlbench and mclust). 

\subsection{Performance across three state of the art regression methods}

Following the same data is put through three high-performance regression method: Generalised Linear Models with Regularisation, Boosting, and Random Forests. In R, they are called "glmnet", "xgboost" and "ranger" respectively. The results show that using correlation as the metric, the synthetically sampled points improve the performance of the regression on the test set between prediction and with-held test set. Furthermore, this is with statistical significance according to a t-test of the correlation between original data and synthetic data, indicated by (*****).

\begin{figure}[H]
    \includegraphics[width=1.0\textwidth]{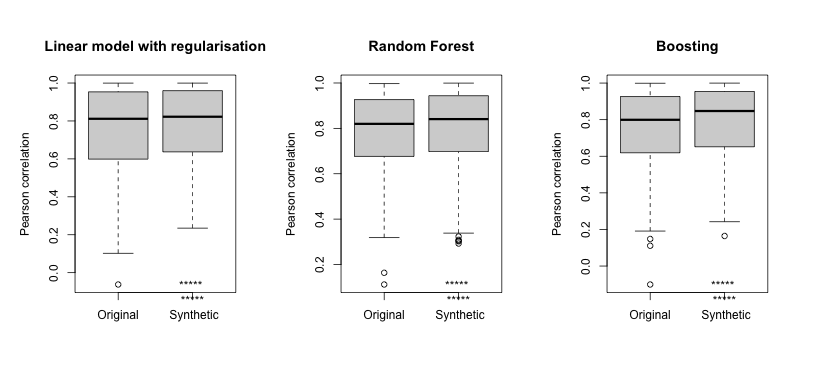}
\end{figure}

\subsection{Similar topology represented between synthetic and original data}

Using the data from Cata et al on prostate cancer [7], the original data contains details on 316 men with prior and follow-up tests conducted on the prostate. Learning from the original data points, several hundred thousand more synthetic samples were generated. 

When the Gleason score, a measure of prostate cancer severity, is shown via the biopsy score (X-axis) and surgical score (Y-axis), we notice the distribution of patient occurrence is similar between synthetic and original. Furthermore, it is apparent in the synthetic dataset which class the synthetic patient falls into; meaning we can re-balance the dataset and pick and equal number of patients from each class based on Gleason score. Evidently, taking the data as is means the data is unbalanced with most patients falling into the Biopsy Gleason Score of 1 (least severity).

\begin{figure}[H]
    \includegraphics[width=1.0\textwidth]{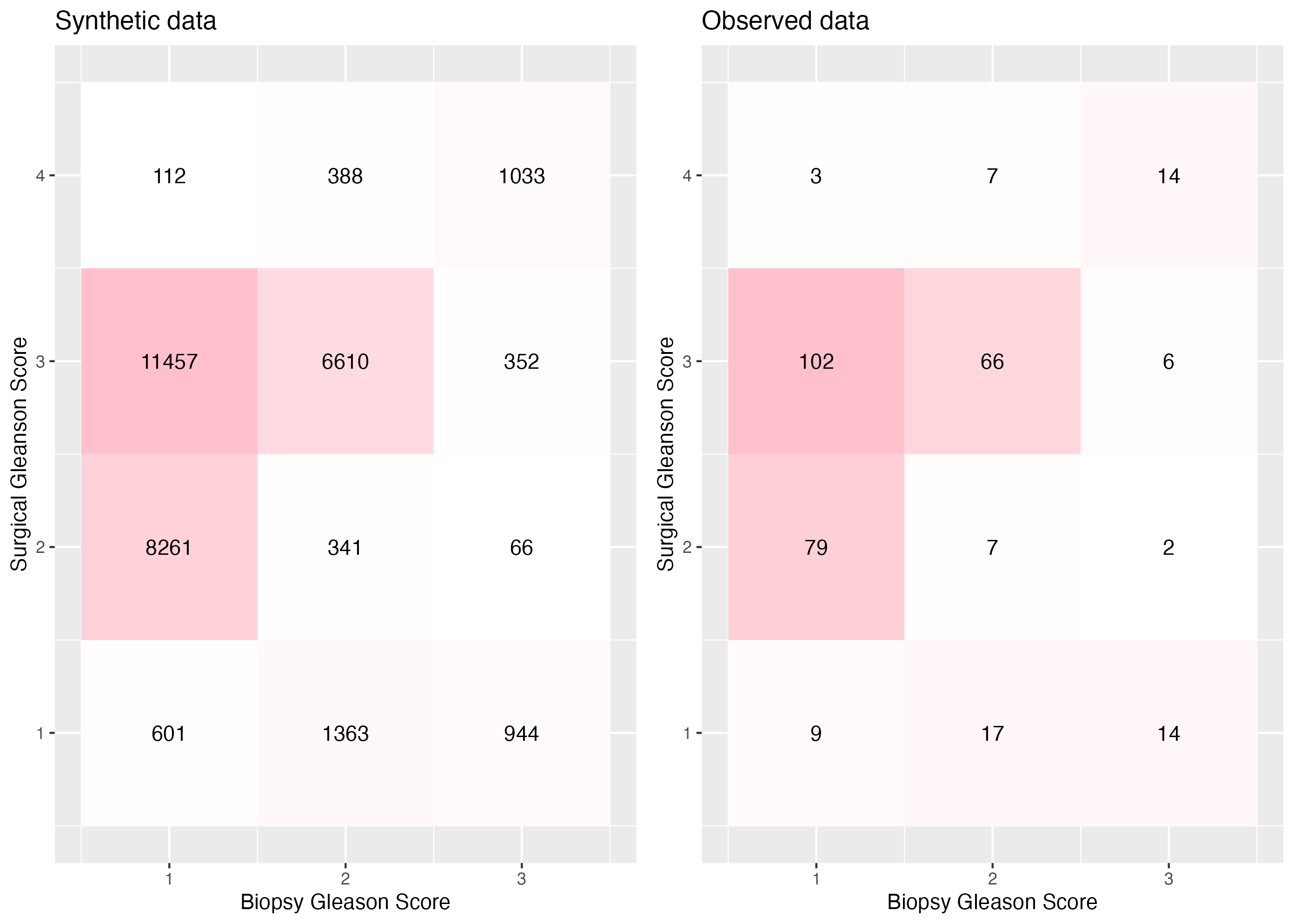}
\end{figure}

\subsection{Biological validation with differential gene tests on real and synthetic data}

Genes are expressed to form proteins. Different parts of the body and different tissues have different functions and thus express differential levels of protein and genes. The greater the difference in function, the more likely the difference in gene expression can be detected. 

The test here is to compare the similarity between synthetic and real RNA-seq on the Brain GTEX dataset (Genotype-Tissue Expression dataset). A standard differential based test on the gene expresion is done, as per the literature. The closer the differential-based scores are per synthetic and real, the more similar the synthetic dataset represents real biological signals (via a correlation analysis).

Here, with a simple correlation calculation, the cerebellum and cerebellar hemisphere of the brain are compared against other regions of the brain in a differential gene test comparing tissues. Notice that there are many other parts of the brain tested in this analysis, but the top two highest correlation scores against all other brain tissues are similar - the cerebellum, which is part of the cerebellar hemisphere. 

\begin{center}
\begin{tabular}{ |c||c|c| } 
    \hline
      & Cerebellum & Cerebellar Hemisphere \\
    \hline\hline
    Cortex & 0.852 & 0.808 \\
    Putamen (basal ganglia) & 0.853 & 0.864 \\
    Hypothalamus & 0.906 & 0.896 \\
    Hippocampus & 0.903 & 0.899 \\
    Substantia nigra & 0.904 & 0.881 \\
    Amygdala & 0.915 & 0.895 \\
 \hline
\end{tabular}
\end{center}

These two brain parts are most distinct compared to the other parts of the brain, and thus present well in a differential gene based test comparing tissue function.

\section{Algorithmic properties}
\subsection{Density Estimation and Mixture Modelling}

A Gaussian mixture model can model any known density estimate. Especially with the numerous literature findings on multivariate gaussian models, mixture modelling and multivariate sampling from existing libraries makes this desirable. 

The structure $A$ provided by extending the sample space via the latent sample parameter $\alpha$ is similar to K-nearest neighbour interpolation. The latent sample space $\alpha$ is treated as a data process from the Gaussian mixture model and the covariance structures based on the latent samples exhibits a graphical structure: variance-covariance matrices can be considered a graph, or in terms of networks analysis, an "adjacency" matrix. 

\subsection{Concepts on Generative Sampling}
For example, take $ samples_{n,g} \in \Omega_{S} $ and let the observed sampling structure be defined by $D$. A sampling structure can be exhibited by a graphical model: a complex network or random graph, a Bayesian network with prior distributions, or quite simply, a (mixture) of multivariate Gaussian models. Here, let $D$ be the variance-covariance structure of the Gaussian (mixture) model $GMM(L_g;\theta_g)$ to represent the sampling structure of $\Omega_{S}$.

Treating samples as structures or collective units rather than an individual set of random samples contains more information for later analysis by other statistical or machine learning models. Thus, as a whole the newly generated samples contains more sample structure than a random set. The aim of any sampling strategy is to gain insights from a structural (parametric) or statistical (probabilistic) analysis of the data (e.g. $D_g \in GMM(L_g;\theta_g(\mu_g, D_g))$ ) to relate to a closer representation of the data generating process (e.g. $\Omega_{S}$).

\section{Conclusion}
With the ability to generate fully synthetic samples without identifiers, data can be shared more easily. In fact, the concerns regarding data privacy can be looked at in light of fully synthetic data that surpasses the original dataset in terms of accuracy and precision. The synthetic samples have been shown to be robust to statistical biological noise and image variation.

\section{References}

[1] Sharyl J. Nass, Laura A. Levit, and Lawrence O. Gostin. Beyond the HIPAA Privacy Rule.
National Academies Press, 2009.

[2] Eric Horvitz and Deirdre Mulligan. Data, privacy, and the greater good. Science, 349(6245):253–
255, 2015.

[3] Priyank Jain, Manasi Gyanchandani, and Nilay Khare. Big data privacy: a technological
perspective and review. Journal of Big Data, 3(1):25, Nov 2016.

[4] Hong-Li Yin and Tze-Yun Leong. A model driven approach to imbalanced data sampling in
medical decision making. Studies in health technology and informatics, 160:856–60, 01 2010.

[5] N. V. Chawla, K. W. Bowyer, L. O. Hall, and W. P. Kegelmeyer. SMOTE: Synthetic minority
over-sampling technique. Journal of Artificial Intelligence Research, 16:321–357, June 2002.

[6] Lindenbaum, O., Stanley, J. S., Wolf, G., Krishnaswamy, S. (2018). Geometry-Based Data Generation (Version 4). arXiv. https://doi.org/10.48550/ARXIV.1802.04927

[7] Cata JP, Klein EA, Hoeltge GA, Dalton JE, Mascha E, O'Hara J, Russell A, Kurz A, Ben-Elihayhu S, Sessler DI. Blood storage duration and biochemical recurrence of cancer after radical prostatectomy. Mayo Clin Proc. 2011 Feb;86(2):120-7. doi: 10.4065/mcp.2010.0313. Erratum in: Mayo Clin Proc. 2011 Apr;86(4):364. PMID: 21282486; PMCID: PMC3031436.

[8] Amit Bermanis, Amir Averbuch, and Ronald R. Coifman. Multiscale data sampling and function
extension. Applied and Computational Harmonic Analysis, 34(1):15–29, January 2013.

[9] Jerome Friedman, Trevor Hastie, and Robert Tibshirani. Regularization paths for generalized
linear models via coordinate descent. Journal of Statistical Software, 33(1), 2010

\end{document}